\documentclass[10pt,journal,compsoc]{IEEEtran}
\ifCLASSOPTIONcompsoc
  \usepackage[nocompress]{cite}
\else
  \usepackage{cite}
\fi
\ifCLASSINFOpdf
\else
\fi
\newcommand{\ie}{\textit{i}.\textit{e}.}
\newcommand{\eg}{\textit{e}.\textit{g}.}
\usepackage[dvipsnames]{xcolor}
\usepackage{graphicx}
\usepackage{enumerate}
\usepackage{amsmath,amssymb}
\usepackage{multirow}
\usepackage{verbatim}
\usepackage{threeparttable}
\usepackage{subfigure}
\usepackage[colorlinks]{hyperref}
\usepackage{color, colortbl}
\usepackage{bbding}
\DeclareMathOperator*{\argmax}{argmax}

\begin{document}
%
\title{Dynamic Feature Regularized Loss for Weakly Supervised Semantic Segmentation}
%
%
%
%

\author{Bingfeng Zhang, 
	Jimin Xiao,~\IEEEmembership{Member,~IEEE,}
	Yao Zhao, ~\IEEEmembership{Senior Member,~IEEE} 
	\thanks{B.~Zhang is with University of Liverpool, UK, and also with School of Advanced Technology, Xi'an Jiaotong-Liverpool University, Suzhou, P.R. China  (e-mail: bingfeng.zhang@liverpool.ac.uk).}
	\thanks{J.~Xiao is with School of Advanced Technology, Xi'an Jiaotong-Liverpool University, Suzhou, P.R. China (e-mail: jimin.xiao@xjtlu.edu.cn). (Corresponding author: Jimin Xiao).}
	\thanks{Y. Zhao is with Institute of Information Science, Beijing Jiaotong University, Beijing, China (e-mail: yzhao@bjtu.edu.cn).}
}

\IEEEtitleabstractindextext{%
\begin{abstract}
We focus on tackling weakly supervised semantic segmentation with scribble-level annotation. The regularized loss has been proven to be an effective solution for this task. However, most existing regularized losses only leverage static shallow features (color, spatial information) to compute the regularized kernel, which limits its final performance since such static shallow features fail to describe pair-wise pixel relationship in complicated cases. In this paper, we propose a new regularized loss which utilizes both shallow and deep features that are dynamically updated in order to aggregate sufficient information to represent the relationship of different pixels. 
Moreover, in order to provide accurate deep features, we adopt vision transformer as the backbone and design a feature consistency head to train the pair-wise feature relationship. Unlike most approaches that adopt multi-stage training strategy with many bells and whistles, our approach can be directly trained in an end-to-end manner, in which the feature consistency head and our regularized loss can benefit from each other. Extensive experiments show that our approach achieves new state-of-the-art performances, outperforming other approaches by a significant margin with more than 6\% mIoU increase. 

\end{abstract}

}

\maketitle

\IEEEdisplaynontitleabstractindextext

%
\IEEEpeerreviewmaketitle

\section{Introduction}
%
%
%
%
\IEEEPARstart{F}{ully} supervised semantic segmentation~\cite{long2015fully,chen2014semantic} has been witnessing great success with dense pixel-level annotation. However, such pixel-level annotation is time-consuming and highly relies on human effort. Weakly supervised semantic segmentation, aiming to make pixel-level prediction under weak supervision signals (scribble~\cite{lin2016scribblesup, tang2018regularized}, bounding box~\cite{kulharia2020box2seg, zhang2021affinity}, point~\cite{bearman2016s, ke2021universal} and image-level~\cite{zhang2019reliability,liu2020leveraging}) is a solution for this problem. In this paper, we focus on weakly supervised semantic segmentation with scribble annotation. The main challenge of this task is how to train the segmentation model with limited supervision.

Most recent state-of-the-art approaches can be divided into two main categories: pseudo-label based approaches~\cite{pu2018graphnet, zhang2021affinity} and loss function based approaches~\cite{tang2018normalized, tang2018regularized, ke2021universal, wang2019boundary}. Pseudo-label based approaches focus on generating more pseudo labels through expanding the initial annotations so that the segmentation model receives more completed pixel-level labels as supervision. But such approaches usually need multi-stage training process with many bells and whistles. For example, in $A^2$GNN \cite{zhang2021affinity},  three different models are used for this task. Loss function based approaches concentrate on directly utilizing limited labels to train the segmentation model with well-designed loss functions. However, some approaches~\cite{ke2021universal, wang2019boundary} rely on extra dataset~\cite{xie2015holistically, arbelaez2010contour} to provide edges or boundaries information as supervision, while some loss function based approaches~\cite{tang2018normalized, tang2018regularized} still need multi-round training procedures. Although Gated CRF loss~\cite{obukhov2019gated} can be directly trained in an end-to-end manner, its performance is limited as it solely relies on static shallow feature (color and spatial information), which fails to capture accurate pair-wise pixel relationship. For example, the shallow features are similar for a pixel pair which belongs to different objects with similar color and close spatial positions (\eg, a white dog close to a white cat). In this case, the shallow features cannot accurately describe the semantic relationship of different pixels. Using such information to compute the  regularized loss enforces the network to be optimized towards an inaccurate direction. More importantly, since shallow features are static, such process can not be corrected in the whole training period. Therefore, it is important to introduce more comprehensive representations for the regularized loss.

In this paper, we propose a new Dynamic Feature Regularized (DFR) loss function in the semantic segmentation head to overcome the aforementioned drawbacks. Our DFR loss makes full use of both static shallow feature and dynamic deep feature, which provides more sufficient information to describe the semantic similarity of different pixels. However, pixel features from the same semantic category may not be sufficient similar, so we design a feature consistency head to enforce this goal. Our feature consistency head utilizes the highly confident prediction from our semantic segmentation head as supervision. It closes up feature distance for pixels from the same semantic category and widens feature distance for pixels from different categories. 

Our semantic segmentation head and feature consistency head are directly coupled as they enhance each other mutually. On one hand, deep feature from our feature consistency head provides a third dimension of input for the regularized loss of the semantic segmentation head, so as to produce accurate semantic prediction. On the other hand, accurate semantic prediction provides more reliable supervision for the feature consistency head, empowering it to build more discriminative features. As a result, compared to solely relying on static shallow feature to compute regularized kernel, the interaction between the two heads allows the deep feature to dynamically change, which also enables deep feature level self-correction and mitigates the negative influence of the inaccurate shallow feature.

Meanwhile, in order to keep high computational efficiency for our loss functions, a local window is used to restrict the loss computing region. Thus, in order to provide more comprehensive information, we adopt vision transformer~\cite{dosovitskiy2020image,liu2021swin} as our backbone since such model can extract global feature representations. 

Our approach can be directly trained in an end-to-end manner and it does not rely on any extra dataset to provide supervision. Without applying any post-processing method such as dense CRF~\cite{krahenbuhl2013parameter} to refine the results, our approach significantly outperforms the previous state-of-the-art approaches, with an mIoU increase of more than 6\%. Our contributions are summarized as follows:

\begin{itemize}	

	\item We propose a new dynamic feature regularized loss for weakly supervised semantic segmentation. Our regularized loss combines both static shallow and dynamic deep features for the regularized kernel, which can better represent the pair-wise pixel relationship. 
	
	\item We design a new feature consistency head to produce consistent features for pixels of same semantic category, enabling to build more accurate pair-wise pixel relationship. Meanwhile, we introduce vision transformer to strengthen the feature representation. To the best of our knowledge, this is the first work that uses transformer architecture for this task.
	
	\item Our approach achieves state-of-the-art performances on PASCAL VOC 2012 (\emph{val}: 82.8\%, \emph{test}: 82.9\%) and PASCAL CONTEXT (\emph{val}: 52.9\%), outperforming other approaches by a large margin (more than 6\% and 12\% mIoU increases on PASCAL VOC 2012 and PASCAL CONTEXT, respectively). 
	
\end{itemize}

\section{Related Works}

\subsection{Fully Supervised Semantic Segmentation}
Fully supervised semantic segmentation has made significant progress with advances in deep neural network especially the fully convolutional network (FCN)~\cite{chen2014semantic}. Deeplab v2~\cite{chen2014semantic} proposed an ASPP module which utilized dilated convolution to increase respective filed while Deeplab v3+~\cite{chen2017rethinking} introduced an encoder-decoder structure to up-sample its prediction. PSPNet~\cite{zhao2017pyramid} designed a pyramid pooling module in an FCN architecture to generate more refined object details. SegSort~\cite{hwang2019segsort} proposed a clustering method to segment objects. Tree-FCN~\cite{song2019learnable} designed a learnable tree filter to utilize the structural property to model long-range dependencies. Recently, inspired by the success of vision transformer~\cite{dosovitskiy2020image} for image classification, some new vision transformer architectures~\cite{liu2021swin, wang2021pyramid} were introduced for fully supervised semantic segmentation, which led to clear performance improvement. Specifically, PVT~\cite{wang2021pyramid} designed a pyramid vision transformer to make dense prediction. Swin-transformer~\cite{liu2021swin} proposed to utilize a local window to improve the attention computing efficiency and a shifted window to extract global information. In this paper, we also introduce vision transformer to provide strengthened feature representations.   

\subsection{Weakly Supervised Semantic Segmentation}
According to the weak supervision signal, weakly supervised semantic segmentation can be divided into scribble-level~\cite{lin2016scribblesup, tang2018regularized, obukhov2019gated}, bounding-box level~\cite{lee2021bbam, oh2021background}, point-level~\cite{bearman2016s} and image-level~\cite{zhang2019reliability,zhang2020causal}. 

For scribble-level setting, ScribbleSup~\cite{lin2016scribblesup} proposed to utilize super pixel~\cite{achanta2012slic} to expand initial annotation and design a loss function to use the expanded supervision. $A^2$GNN~\cite{zhang2021affinity} designed a graph-based approach to generate pseudo labels from scribble supervision, which is then used to train a segmentation model. Tang \emph{et.al.}~\cite{tang2018normalized, tang2018regularized} proposed Normalized Cut loss and Kernel Cut loss to directly use initial labels as supervision. However, both Normalized Cut and Kernel Cut need multi-round training. Gated CRF loss~\cite{obukhov2019gated} improves the efficiency of Kernel Cut loss through adding a gate operation. However, only relying static shallow feature cannot build accurate relationship for different pixels. SPML~\cite{ke2021universal} used SegSort~\cite{hwang2019segsort} as the backbone and HED contour detector~\cite{arbelaez2010contour} as extra supervision. BPG~\cite{wang2019boundary} designed an iterative strategy to produce the fine-grained feature maps, which also applied contour detector~\cite{xie2015holistically, arbelaez2010contour} to provide boundary supervision. In this paper, we propose a new regularized loss which does not need extra supervision and our approach can be directly trained in an end-to-end manner.

\section{Methodology}
\subsection{Overview}
\begin{figure*}
\centering
\includegraphics[width=0.92\textwidth]{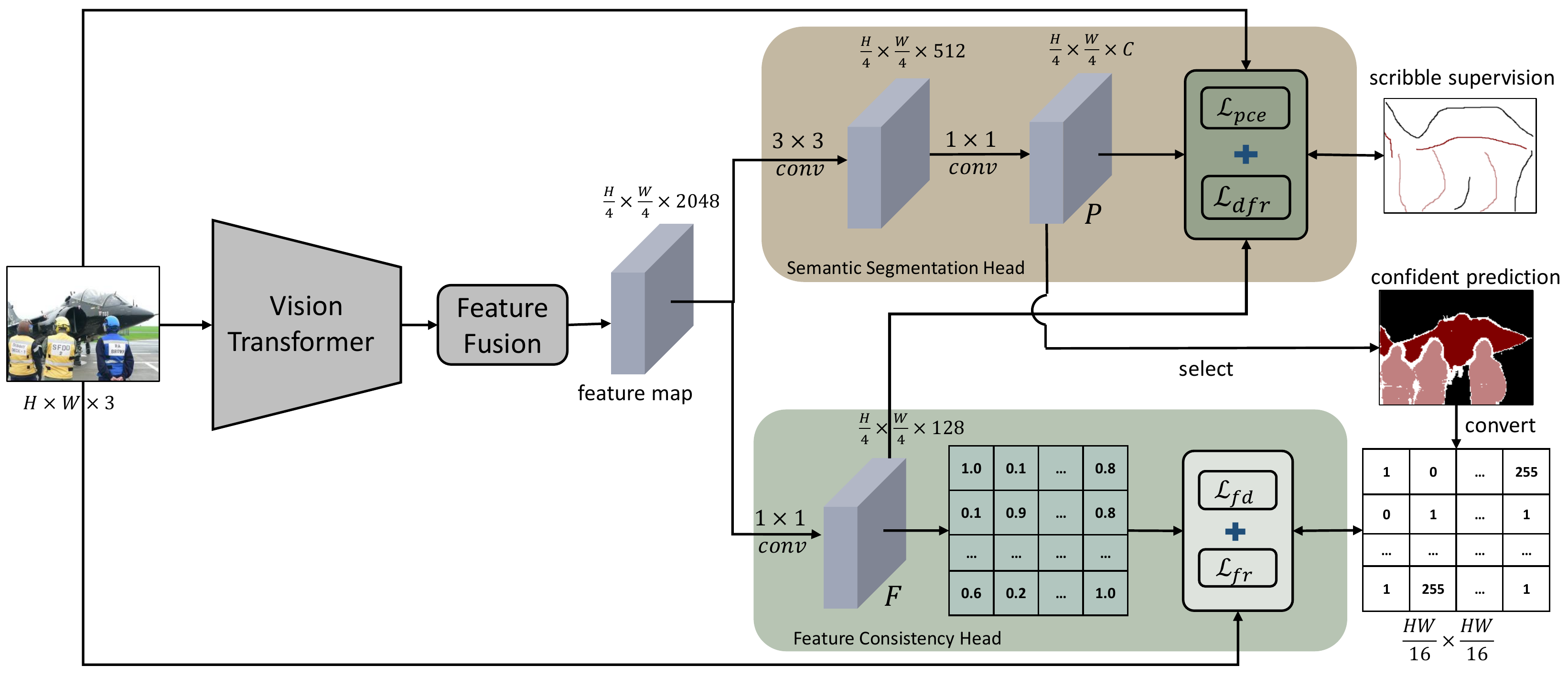}
\caption{The framework of our proposed approach. Firstly, an image is input to the vision transformer to generate its feature maps, then the feature maps from all blocks are fused to generate a shared feature map, which is input to both the semantic segmentation head and the feature consistency head. The semantic segmentation head is used to make semantic prediction and provide highly confident regions as pseudo labels for the feature consistency head. Meanwhile, the feature consistency head is used to produce consistent features for pixels with the same semantic category, which are in turn used in the regularized loss of the semantic segmentation head. Note that both the semantic segmentation head and feature consistency head are used during training while only the semantic segmentation head is used during inference.}
\label{fig:framework}
\end{figure*}

Fig.~\ref{fig:framework} shows the overall framework of our approach. First, we use vision transformer as the backbone to generate the feature maps.  Then the feature maps are input to the feature fusion module to generate the shared feature map for the semantic segmentation head and feature consistency head. The semantic segmentation head is to make semantic prediction and provide supervision for the feature consistency head. The feature consistency head enforces feature consistency for pixels with the same semantic category based on the supervision from the semantic segmentation head, which in turn provides reliable dynamic feature for the semantic segmentation head. 

The semantic segmentation head utilizes two loss functions: partial cross-entropy loss and our proposed dynamic feature regularized loss. Partial cross-entropy loss uses the scribble-annotation as supervision while our proposed dynamic feature regularized loss applies the original image information and feature map from the feature consistency head to produce regularized kernel.

The feature consistency head also introduces two loss functions: feature distance loss and feature regularized loss. Feature distance loss uses the predicted highly confident pseudo labels from the semantic segmentation head as supervision. Feature regularized loss solely uses the shallow feature as the kernel to compute feature distance. 

The whole framework is trained in an end-to-end manner, the loss function is defined as:

\begin{equation}
	\mathcal{L}=\underbrace{\mathcal{L}_{\text{pce}}+\lambda_1\mathcal{L}_{\text{dfr}}}_{\text{semantic  head}}+\lambda_2(\underbrace{\mathcal{L}_{\text{fd}}+\mathcal{L}_{\text{fr}}}_{\text{feature head}}),\label{eq:loss_all}
\end{equation}
where $\lambda_1$ and $\lambda_2$ are loss weights. $\mathcal{L}_{\text{pce}} \text{ and }\mathcal{L}_{\text{dfr}}$ are the loss functions for the semantic segmentation head. $\mathcal{L}_{\text{pce}}$ is the partial cross-entropy loss, which uses the scribble annotation as supervision. $\mathcal{L}_{\text{dfr}}$ is our proposed regularized loss. Both of them will be introduced in Sect.~\ref{sec:Reg}. $\mathcal{L}_{\text{fd}} \text{ and }\mathcal{L}_{\text{fr}}$ are the loss functions for the feature consistency head, $\mathcal{L}_{\text{fd}}$ is the feature distance loss, which uses the prediction of the semantic segmentation head as supervision. $\mathcal{L}_{\text{fr}}$ is the feature regularized loss. Both $\mathcal{L}_{\text{fd}} \text{ and }\mathcal{L}_{\text{fr}}$ will be introduced in Sect.~\ref{sec:FeatureHead}. 
 
\subsection{Semantic Segmentation Head}\label{sec:Reg}
The semantic segmentation head is to make semantic prediction, which includes several convolution layers to produce the final probability map $P$, a partial cross-entropy loss to utilize the scribble annotation and our proposed dynamic feature regularized loss to restrict the prediction of the whole map. 

Specifically, the partial cross-entropy loss is:
\begin{equation}
    \mathcal{L}_{\text{pce}} = -\frac{1}{N_s}\sum_{i=1}^{hw}[M_s(i) \neq 255]log(P^{t}(i)),
\end{equation}
where $P^{t}(i)$ is the probability of pixel $i$ to be classified to the ground truth class. $N_{s}$ is the annotated pixel number. $h$ and $w$ correspond to the height and width of the feature map, respectively. $M_s$ is the provided scribble annotation and 255 means that there is no annotation. $\left [ \cdot \right ]$ is the Iverson bracket operation, which equals to 1 if the inside condition is true, otherwise it equals to 0. 

For scribble annotation, the main limitation is that very few pixel-level labels are provided, \eg, 3\% pixels are annotated in PASCAL VOC 2012 dataset~\cite{tang2018regularized}. In this case, using partial cross-entropy loss is not enough. Therefore, we design a new DFR loss to impose restriction on the prediction of the model. Our intuition is that for two different pixels $i$ and $j$, if their features are highly similar, the probability for them to belong to the same category is high. 

In order to impose the above restriction and keep high computing efficiency, for two different pixels $i$ and $j$, we only compute the loss when both of them locate within a local window:
\begin{equation}
	R = \left\{ (i,j) \bigg| \left|i_x -j_x\right| \leqslant  r \text{ and } \left|i_y -j_y\right| \leqslant r \right\}, \label{eq:Rij}
\end{equation}
where $R$ is the effective pixel pair set. $i_x$ and $j_x$ represent the x-coordinate, $i_y$ and $j_y$ represent the y-coordinate. $r$ is the window size.

Then our proposed DFR loss is:
\begin{equation}
	\mathcal{L}_{\text{dfr}} = \frac{1}{hw}\sum_{i=1}^{hw}\sum_{(i,j)\in R}\left [ i\neq j\right ]\varphi (i,j),\label{eq:Lreg}
\end{equation}
where $\varphi (i,j)$ is the loss for pixels $i$ and $j$, which follows the definition:
\begin{equation}
\begin{aligned}
\varphi (i,j) &= \sum_{c \in C}\sum_{c' \in C}\left [ c\neq c'\right ]K(i,j)P^{c}(i)P^{c'}(j)\\
&=K(i,j)\sum_{c \in C}P^{c}(i)\left(1-P^{c}(j)\right)\\
&=K(i,j)\left(1-\sum_{c \in C}P^{c}(i)P^{c}(j)\right)\label{eq:varphi(i,j)},
\end{aligned}
\end{equation}
where $C$ is the class set, \ie, $C=\left\{c_1,c_2,...,c_N\right\}$. $P^{c}(i)$ and $P^{c'}(j)$ are the probabilities for pixels $i$ and $j$ to be classified to class $c$ and $c'$, respectively, which are provided by the network (after softmax layer). $K(i,j)$ is the regularized kernel, which is defined as a Gaussian kernel:
\begin{equation}
\resizebox{0.9\columnwidth}{!}{$
K(i,j) = exp\left(-\frac{\left \| S_i-S_j\right \|^2}{2\sigma_{1}^2}-\frac{\left \| I_i-I_j\right \|^2}{2\sigma_{2}^2}-\frac{\left \| F_i-F_j\right \|^2}{2\sigma_{3}^2}\right),
$}\label{eq:Kij}
\end{equation}
where $||\cdot||^2$ is the L2 distance. $S_i$ and $S_j$ correspond to the pixel positions for pixel $i$ and $j$. $I_i$ and $I_j$ are the RGB information of pixel $i$ and $j$. $F_i$ and $F_j$ are the deep features of pixels $i$ and $j$ from the feature consistency head. 

The previous regularized loss functions~\cite{tang2018normalized, tang2018regularized,obukhov2019gated} only adopt the position and RGB information to compute the kernel. However, both position and RGB information in Eq.~(\ref{eq:Kij}) are static, once the two types of features fail to correctly describe the true relationship of a pixel pair, the network will be optimized towards an inaccurate direction, and such a problem cannot be addressed during the whole training period. 

Different from the previous approaches, we introduce the dynamic deep feature, which is provided by the feature consistency head (as described in Sect.~\ref{sec:FeatureHead}), to compute the regularized kernel. Note that when using deep features to compute the regularized kernel, they are regarded as non-gradient values. Through introducing dynamic feature to compute regularized kernel, on  one hand, more comprehensive representation for pixel relationship is provided. On the other hand, dynamic features allow the network to correct its previous results. The remaining task is how to guarantee deep features accurately representing the relationship of different pixels, which is addressed in Sect.~\ref{sec:FeatureHead}.

\subsection{Feature Consistency Head}\label{sec:FeatureHead}
In order to provide correct relationship for deep features of different pixels, we design a feature consistency head. Our motivation is that for two pixels $i$ and $j$, if they belong to the same class, their features should have high similarity. If they belong to different classes, the similarity of their features should be low.

Based on above analysis, we  need to provide supervision for the feature relationship.  We select the predicted labels with highly confident scores from the semantic segmentation head as supervision:
\begin{equation}
M(i) = 
\begin{cases}
\argmax\limits_{c\in C}(P^c(i)),  & \max\limits_{c\in C}(P^c(i)) > \gamma \\
255, & \text{else}
\end{cases}, \label{eq:M_i}
\end{equation}
where $M(i)$ is the semantic label for pixel $i$, $255$ means that it is not annotated to any class. $P^c(i)$ is the predicted probability for class $c$.

Then the supervision is converted to the pair-wise pixel relationship. Following the operation in Sect.~\ref{sec:Reg}, we use the same local window to restrict the computing region. Considering that some pixels are not annotated, so the effective pixel pairs are:
\begin{equation}
\begin{aligned}
R_A=\left\{(i,j)|M(i) \neq 255 \text{ and } M(j) \neq 255  
\right.  \\
\text{and }  \left. (i,j) \in R \right\}, \label{eq:RA}
\end{aligned}
\end{equation}
where $R_A$ is the effective pixel pair set. $R$ is the set defined in Eq.~(\ref{eq:Rij}). After that, the supervision $M$ is converted to the pair-wise pixel relationship label:    
\begin{equation}
	A(i,j)=\left\{\begin{matrix}
	1, &  (i,j) \in R_A \text{ and } M(i) = M(j) \\  
	0, &  (i,j) \in R_A \text{ and } M(i) \neq M(j) \\ 
	255, & else
	\end{matrix}\right.\label{eq:Aij}.
\end{equation}

Eq.~(\ref{eq:Aij}) indicates that when pixels $i$ and $j$ belong to the same class, they have strong relationship (set as $1$). If they belong to different classes, they should have weak relationship (set as $0$). $255$ means the pixel pair is ignored. In order to utilize such supervision, we compute the feature distance as the feature relationship for the pixel pair in $R_A$:
\begin{equation}
	D(i,j) = exp\left(-\frac{||F_i-F_j||}{d}\right),\label{eq:Dij}
\end{equation} 
where $||\cdot||$ is L1 distance. $d$ is the channel dimension of the feature map. Both $F_i$ and $F_j$ are the final features of pixels $i$ and $j$ from the feature consistency head.

Finally, the feature distance loss is:
\begin{equation}
\begin{aligned}
\mathcal{L}_{\text{fd}}=&-\frac{1}{|A_{\text{bg}}^+|}\sum\limits_{(i,j) \in A_{\text{bg}}^+ }A(i,j)log(D(i,j))\\
&-\frac{1}{|A_{\text{fg}}^+|}\sum\limits_{(i,j) \in A_{\text{fg}}^+ }A(i,j)log(D(i,j))\\ 
&-\frac{2}{|A^-|}\sum\limits_{(i,j) \in A^-}(1-A(i,j))log(1-D(i,j)),\label{Lfd} 
\end{aligned}
\end{equation} 
where $A_{\text{bg}}^+$ is the pixel pair set that $A(i,j)=1$ and the label of $i$ and $j$ is background. $A_{\text{fg}}^+$ is the pixel pair set that $A(i,j)=1$ and the label of $i$ and $j$ is foreground. $A^-$ corresponds to the pixel pair set that $A(i,j)=0$. $|\cdot|$ indicates the number of elements in a set.

Following the same strategy in our semantic segmentation head, we also introduce feature regularized loss since the supervision only provide limited annotations. The feature regularized loss is defined as:
\begin{equation}
	\mathcal{L}_{\text{fr}}= \frac{1}{hw}\sum_{i=1}^{hw}\sum_{(i,j)\in R_A}[i\neq j]K_{\text{f}}(i,j)\left(\frac{||F_i-F_j||}{d}\right),\label{Lfr}
\end{equation}
where $K_{\text{f}}(i,j)$ have the similar formation with Eq.~(\ref{eq:Kij}):
\begin{equation}
	K_{\text{f}}(i,j) = exp\left(-\frac{\left \| S_i-S_j\right \|^2}{2\sigma_{1}^2}-\frac{\left \| I_i-I_j\right \|^2}{2\sigma_{2}^2}\right).\label{eq:Kf}
\end{equation}

From Sect.~\ref{sec:Reg} and Sect.~\ref{sec:FeatureHead}, it can be found that both the semantic segmentation and feature consistency heads receive the online updated information. Specifically, the semantic segmentation head receives the dynamically updated feature from the feature consistency head, while the feature consistency head receives the updated supervision from the semantic segmentation head. On one hand, better supervision enables the feature consistency head to provide more accurate feature relationship. On the other hand, more accurate feature relationship facilitates to produce better semantic segmentation.  Thus, we
argue that with such an interaction mechanism two heads benefit from each other and the final performance is boosted accordingly.

\begin{figure}[t]
	\centering
	\includegraphics[width=\columnwidth]{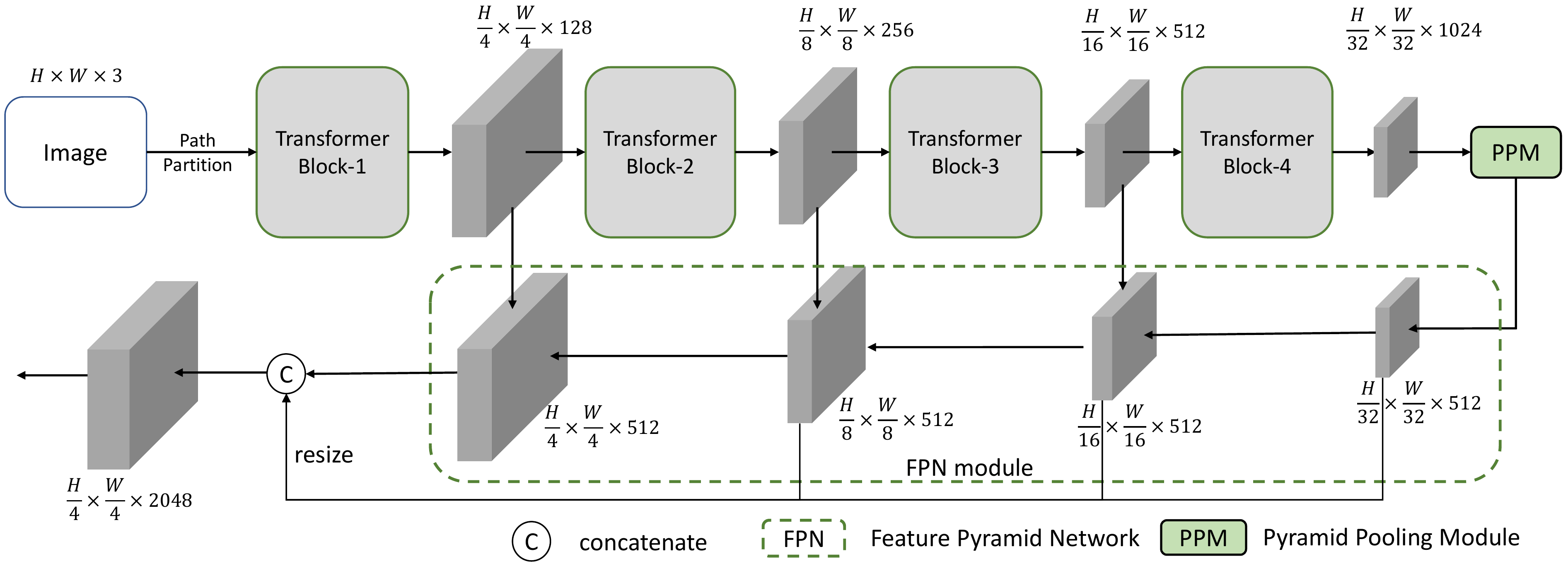}
	\caption{Details of the backbone and the feature fusion process. To fuse features from all stages, we use the same architecture with Swin-transformer~\cite{liu2021swin} and UperNet~\cite{xiao2018unified}, both of which use Pyramid Pooling Module (PPM)~\cite{zhao2017pyramid} and Feature Pyramid Network (PPN)~\cite{lin2017feature} to fuse feature maps.}
	\label{fig:backbone}
\end{figure}

\section{Experiment}
\subsection{Datasets and Evaluation Metric}
We evaluate our method on PASCAL VOC 2012~\cite{everingham2011pascal} and PASCAL CONTEXT~\cite{mottaghi_cvpr14} dataset. For PASCAL VOC 2012 dataset, following the previous approaches~\cite{tang2018normalized, tang2018regularized,obukhov2019gated, ke2021universal} in weakly supervised semantic segmentation, the augmented data SBD~\cite{6126343} is also used and the whole dataset contains 10,582 images for training,  1,449 images for validating and 1,456 images for testing with 20 foreground classes. For PASCAL CONTEXT dataset, it includes 4,998 images for training and 5,105 images for validating with 59 foreground categories. For the scribble annotation, we also follow the previous approaches~\cite{tang2018normalized, tang2018regularized,obukhov2019gated, ke2021universal} to use the supervision provided by ScribbleSup~\cite{lin2016scribblesup}. 
Mean Intersection over Union (mIoU) is adopted as the evaluation metric.
\begin{table*}[]
	\centering
	\caption{Comparison with other state-of-the-art on PASCAL VOC 2012 dataset. Pub.: Publication. Sup.: Supervision. F: Fully-supervised. B: bounding-box level supervision. I: Image-level supervision. S: scribble-level. ``ss" means single scale inference. ``ms" means multi-scale inference. Multi-scale inference is used without explicit indication.}
	\begin{threeparttable}
		\begin{tabular}{lcccccccc}
			\hline
			\multirow{2}{*}{Method} & \multirow{2}{*}{Pub.} & \multirow{2}{*}{Sup.} & \multirow{2}{*}{Backbone} & \multirow{2}{*}{Single-stage} & \multirow{2}{*}{Extra Data} & \multirow{2}{*}{CRF} & \multicolumn{2}{l}{mIoU(\%)} \\ \cline{8-9} 
			& & &  &  &  &  & \emph{val}  & \emph{test}  \\ \hline
			(1) Deeplab-v2~\cite{chen2017deeplab} & TPAMI'18 & F & vgg16 & \checkmark & -  &\checkmark & 71.5 & 72.6\\
			(2) DeepLab-v2~\cite{chen2017deeplab} & TPAMI'18 & F & resnet101  & \checkmark& -&\checkmark & 76.8 & 79.7  \\
			(3) Deeplab-v3+~\cite{chen2018encoder} & ECCV'18  & F & resnet18 &\checkmark& -  &- & 76.7  & - \\
			(4) SegSort~\cite{hwang2019segsort} & ICCV'19& F & resnet101& \checkmark & - &-& 77.3  & -\\
			(5) Tree-FCN~\cite{song2019learnable}  & NeurIPS'19  & F & resnet101 & \checkmark& - & -& 82.3 & - \\
			(6) Swin-Base (ss)~\cite{liu2021swin}\tnote{*}& - & F& transformer& \checkmark& -& -& 82.9& 82.9\\ 
			(7) Swin-Base (ms)~\cite{liu2021swin}\tnote{*}& - & F& transformer& \checkmark& -& -& 84.6&84.4\\\hline
			Box2Seg~\cite{kulharia2020box2seg}& ECCV'20& B & UperNet~\cite{xiao2018unified}&-&-&\checkmark& 76.4 & -  \\
			BAP~\cite{oh2021background} & CVPR'21& B& (2) &-&-&\checkmark& 74.6 & 76.1 \\
			ILLD~\cite{liu2020leveraging}& TPAMI'20& I& Res2Net~\cite{gao2019res2net} &\checkmark&-&\checkmark& 69.4 & 70.4 \\
			AdvCAM~\cite{lee2021anti} & CVPR'21 & I & (2)  &-& -&\checkmark& 68.1  & 68.0 \\
			EDAM~\cite{wu2021embedded}  & CVPR'21 & I  & (2)  &-&\checkmark&\checkmark& 70.9 & 70.6  \\ \hline
			ScribbleSup~\cite{lin2016scribblesup} & CVPR'16  & S & (1)  &-& -   &\checkmark& 63.1  & - \\
			RAWKS~\cite{vernaza2017learning}& CVPR'17 & S & resnet101 &\checkmark&\checkmark  &\checkmark& 61.4& -\\
			NormalizedCut~\cite{tang2018normalized} & CVPR'18 & S & (2) &-& - &\checkmark& 74.5  & -  \\
			GraphNet~\cite{pu2018graphnet}& ACMM'18& S& (2) &-& - &\checkmark& 73.0 & - \\
			KernelCut~\cite{tang2018regularized} & ECCV'18  & S & (2) &-& - &-& 73.0 & - \\
			KernelCut+CRF~\cite{tang2018regularized} & ECCV'18  & S & (2) &-& - &\checkmark& 75.0 & - \\
			GatedCRF~\cite{obukhov2019gated} & NeurIPS'19 & S& (3)  &\checkmark& - &-& 75.5  & -\\
			BPG~\cite{wang2019boundary} & IJCAI'19  & S & (2) &\checkmark&\checkmark& -& 73.2 & - \\
			BPG+CRF~\cite{wang2019boundary} & IJCAI'19  & S & (2) &\checkmark& \checkmark &\checkmark& 76.0 & - \\
			SPML~\cite{ke2021universal} & ICLR'21 & S  & (4)  &-&\checkmark&-& 74.2  & - \\
			SPML+CRF~\cite{ke2021universal} & ICLR'21 & S  & (4)  &-& \checkmark &\checkmark& 76.1  & - \\
			$A^2$GNN~\cite{zhang2021affinity} & TPAMI'21& S & (5) &-& -  &\checkmark& 76.2 & 76.1 \\ \hline
			\emph{DFR-ours (ss)} & -  & S  & (6) &\checkmark&-&-& 81.5 & 82.1\\
			\emph{DFR-ours (ms)} & -  & S  & (7) &\checkmark&-&-& \textbf{82.8} & \textbf{82.9}\\ \hline
		\end{tabular}\label{tab:state-of-the-art}
		\begin{tablenotes}
			\item[*] Reproduced by ourselves.
		\end{tablenotes}
	\end{threeparttable}
\end{table*}

\subsection{Implementation Details}
Our approach mainly includes three network modules: the backbone, the semantic segmentation head and the feature consistency head. For the backbone, we choose Swin-Transformer-Base~\cite{liu2021swin} (with UperNet head~\cite{xiao2018unified} to fuse the features from 4 stages). The details can be found in Fig.~\ref{fig:backbone}. After passing the backbone and the feature fusion stage,  a fused feature map with a dimension of $2048$ is generated. For the semantic segmentation head, we also use the same setting as in Swin-Transformer-Base~\cite{liu2021swin}, which uses the scene head in~\cite{xiao2018unified}. For the feature consistency head, we utilize a $1\times1$ convolutional layer followed by a ReLU function to produce the final feature, and the dimension $d$ of the feature in this head is set as $128$. In Eq.~(\ref{eq:loss_all}), $\lambda_1$, $\lambda_2$ are set as $1\times 10^{-2}$ and $1\times 10^{-3}$, respectively. The window size for Eq.~(\ref{eq:Rij}) and Eq.~(\ref{eq:RA}) is set as 5. $\gamma$ in Eq.~(\ref{eq:M_i}) is $0.98$. $\sigma_1$ and $\sigma_2$ are shared for Eq.~(\ref{eq:Kij}) and Eq.~(\ref{eq:Kf}). $\sigma_1$, $\sigma_2$ and $\sigma_3$ are set as $6$, $0.5$ and $50$, respectively. Note that the RGB is normalized before inputting to the network to compute the kernel.

We use the weights pretrained on ImageNet-22K~\cite{deng2009imagenet} to initialize the model of Swin-Transformer-Base~\cite{liu2021swin}. AdamW~\cite{kingma2014adam} is used as the optimizer with an initial learning rate of $3\times 10^{-5}$ and weight decay of $0.01$. Models are trained on 8 Nvidia Tesla V100 GPUs with batch size of 16 for 40K iterations. During training, we adopt the default settings in mmseg~\cite{contributors2020mmsegmentation}, including random flipping, random rescaling (range is $\left[0.5, 2.0\right]$) and random photometric distortion. The input size is $512\times 512$. During inference, the feature consistency head is not used and multi-scale strategy is used with resolution ratios of $ \{ 0.5,0.75,1.0,1.25,1.5,1.75 \}$.
Other settings follow that in Swin-Transformer-Base~\cite{liu2021swin}. 
 
\subsection{Comparison with State-of-the-Art}
\begin{figure}[t]
	\centering
	\includegraphics[width=0.9\columnwidth]{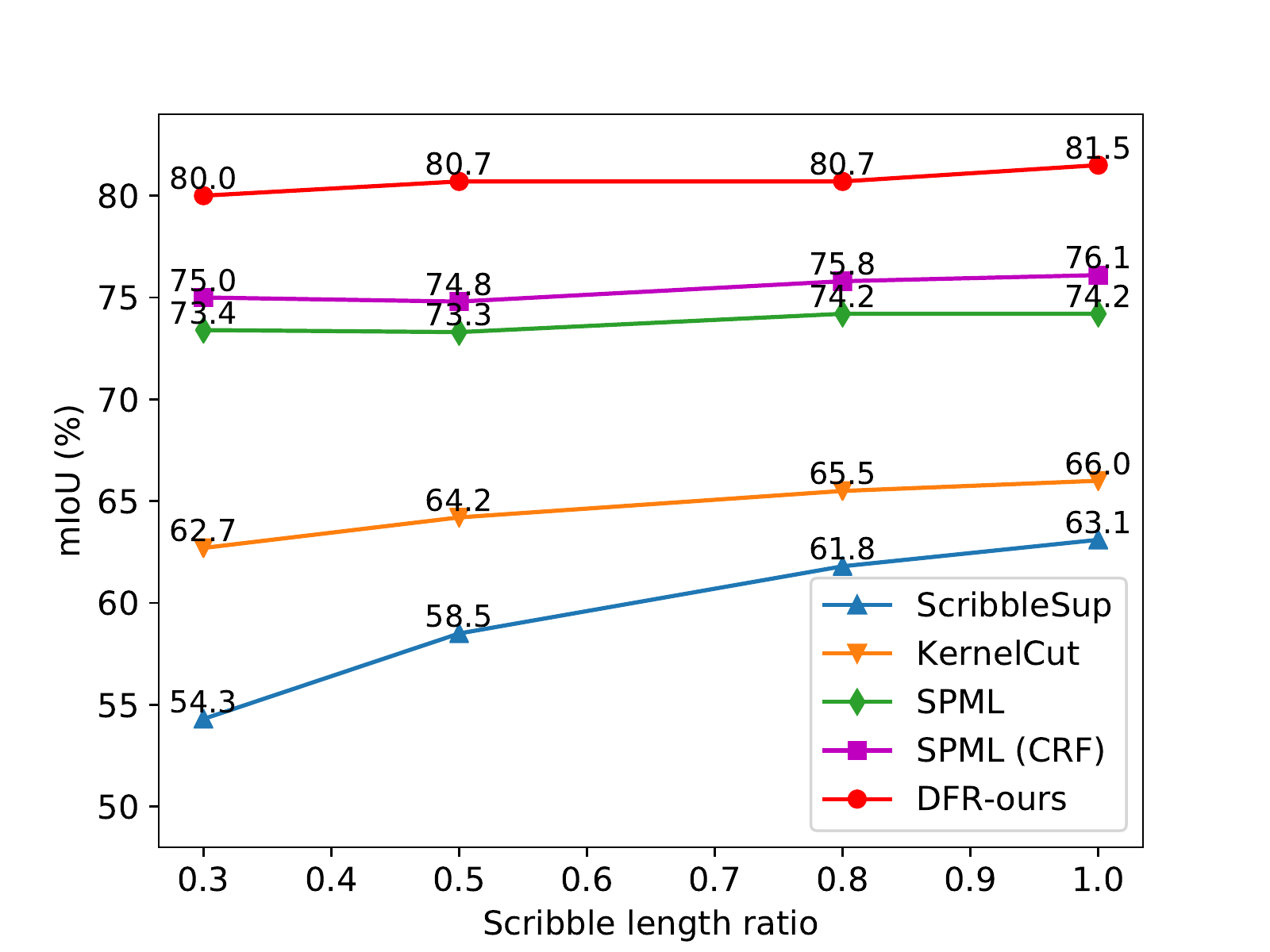}
	\caption{Comparison with other state-of-the-art approaches on PASCAL VOC 2012 \emph{val} set for different scribble lengths.}
	\label{fig:scribble_length}
\end{figure}

In Table~\ref{tab:state-of-the-art}, we compare our approach with other approaches on PASCAL VOC 2012 dataset.  It can be seen that our approach significantly outperforms other approaches. Specifically, $A^2$GNN~\cite{zhang2021affinity} achieves 76.2\% mIoU with dense CRF~\cite{krahenbuhl2013parameter} as post-processing, while we achieve 82.8\% mIoU without using CRF, which brings 6.6\% mIoU gain. Note that $A^2$GNN~\cite{zhang2021affinity} is a multi-stages method which uses more than three individual networks during training, while our approach is a single-stage method. Besides, for the single-stage method, BPG~\cite{wang2019boundary} achieves the best performance, but it used extra dataset (HED contour detector~\cite{xie2015holistically}, pretrained on BSDS500 dataset~\cite{arbelaez2010contour}) to provide edge supervision. We do not rely on any extra dataset and outperform it by 9.6\% mIoU without CRF (82.8\% \emph{v.s.} 73.2\%). SPML used the same extra dataset as BPG~\cite{wang2019boundary} with multi-round training process, and we also significantly outperform it (82.8\% \emph{v.s.} 76.1\%). More importantly, our approach reaches 98.3\% of the upper-bound performance (the fully-supervised case for the single scale setting), showing its effectiveness for this task. It can also be found that using multi-scale strategy brings 1.3\% mIoU increase. For the \emph{test} set, our approach outperforms $A^2$GNN with a clear gain of 6.8\%. Generally, without using any extra-dataset and post-processing, our approach outperforms other approaches by a large margin through single-stage training.

\begin{table*}[]
	\caption{Per-class comparison between our approach and others on PASCAL VOC 2012 \emph{val} set.}
	\resizebox{1\textwidth}{!}{$
		\begin{tabular}{lcccccccccccccccccccccc}
		\hline
		Method  & bkg & aero & bike & bird & boat & bottle & bus & car & cat & chair & cow & table & dog & horse & mbike & person & plant & sheep & sofa & train         & tv & Mean \\ \hline
		KernelCut~\cite{tang2018regularized} & -  & 86.2 & 37.3 & 85.5 & 69.4 & 77.8  & 91.7 & 85.1 & 91.2 & 38.8 & 85.1 & 55.5 & 85.6 & 85.8 & 81.7  & 84.1 & 61.4 & 84.3 & 43.1 & 81.4 & 74.2 & 75.0 \\
		BPG~\cite{wang2019boundary} & 93.4 & 84.8 & 38.4 & 84.6 & 65.5 & 78.8 & 91.4 & 85.9 & 89.5 & 41.0 & 87.3 & 58.3 & 84.1 & 85.2 & 83.7 & 83.6 & 64.9 & 88.3 & 46.0 & 86.3 & 73.9 & 76.0 \\
		SPML~\cite{ke2021universal} & - & 89.0 & 38.4 & 86.0 & 72.6 & 77.9 & 90.0 & 83.9 & 91.0 & 40.0 & 88.3 & 57.7 & 87.7 & 82.8 & 79.1 & 86.5 & 57.1 & 87.4 & 50.5 & 81.2 & 76.9 & 76.1 \\ \hline
		\emph{DFR-ours (ss)} & \textbf{95.0} & \textbf{90.8} & \textbf{39.0} & \textbf{89.8} & \textbf{76.4} & \textbf{82.9} & \textbf{93.8} & \textbf{87.3} & \textbf{94.9} & \textbf{49.4} & \textbf{92.7} & \textbf{66.2} & \textbf{90.9} & \textbf{89.9} & \textbf{86.8} & \textbf{87.8} & \textbf{71.8} & \textbf{90.4} & \textbf{64.0} & \textbf{92.4} & \textbf{79.4} & \textbf{81.5} \\ \hline
		\end{tabular}\label{tab:per_class}
		$}
\end{table*}

In Table~\ref{tab:per_class}, we report the per-class results on PASCAL VOC 2012 \emph{val} set. It can be seen that our approach generates new state-of-the-art performances for each class. Note that we do not use dense CRF while the other reported approaches use dense CRF as post-processing.

In Fig.~\ref{fig:scribble_length}, segmentation performance comparisons with different scribble lengths are reported. Our approach consistently outperforms other approaches using different scribble lengths. Even only provided with 30\% of original scribble length, our approach still obtains an mIoU of 80.0\%.

\begin{table}[]
	\centering
	\caption{Comparison with other state-of-the-art on PASCAL CONTEXT dataset.}
	\begin{tabular}{lcccc}
		\hline
		Method       & Pub. & Sup.& CRF & mIoU (\%) \\ \hline
		ScribbleSup~\cite{lin2016scribblesup} & CVPR'16&S&\checkmark& 36.1 \\
		RAWKS~\cite{vernaza2017learning} &CVPR'17&S& \checkmark& 37.4 \\
		GraphNet~\cite{pu2018graphnet} &ACMM'18&S& -& 33.9 \\
		GraphNet+CRF~\cite{pu2018graphnet} & ACMM'18&S& \checkmark& 40.2 \\ \hline
		\emph{DFR-ours (ss)} & -&S& -&50.9\\
		\emph{DFR-ours (ms)}& -&S& - &\textbf{52.9}\\ \hline
	\end{tabular}\label{tab:pascal_context}
\end{table}

In Table~\ref{tab:pascal_context}, we compare our approach with others on the PASCAL CONTEXT dataset, it can be seen that our approach also achieves a new state-of-the-art performance, with an mIoU gain of 12.7\%.

\begin{figure*}[t]
	\centering
	\includegraphics[width=\textwidth]{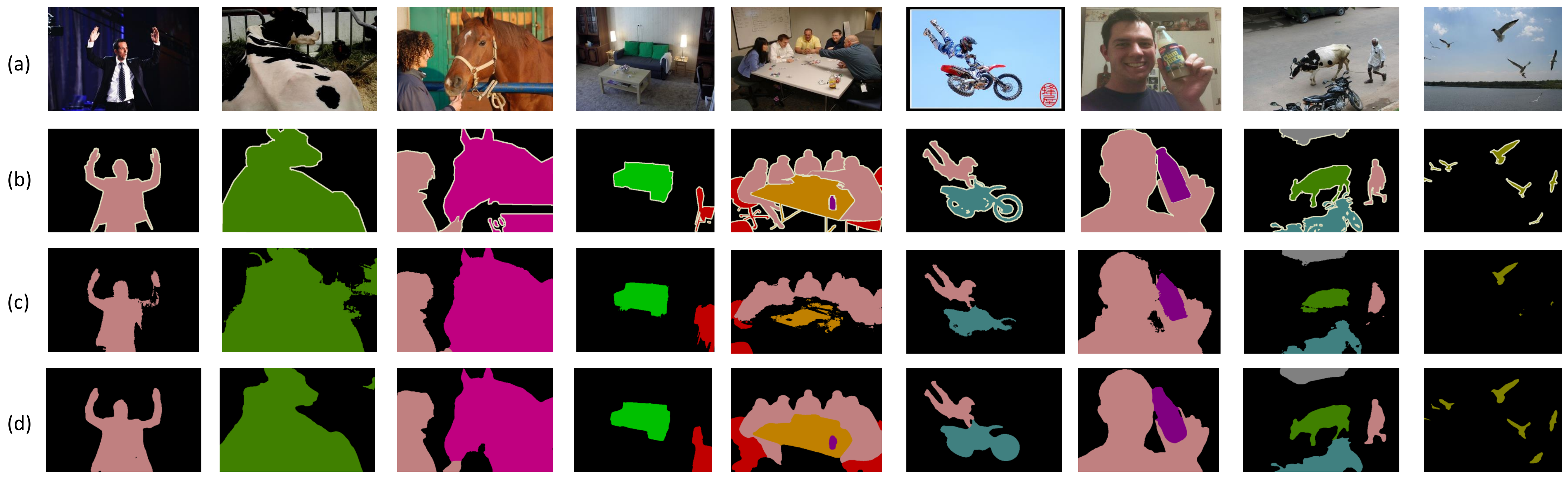}
	\caption{Qualitative comparison between our method and other state-of-the-art approaches on PASCAL VOC 2012 \emph{val} dataset. (a) Original image (b) Ground-truth (c) Results of $A^2$GNN~\cite{zhang2021affinity} with dense CRF~\cite{krahenbuhl2013parameter} as post-processing (d) Our results.}
	\label{fig:vis}
\end{figure*}

In Fig.~\ref{fig:vis}, we show qualitative comparisons between our approach and the previous state-of-the-art approaches. It can be seen that our approach keeps more details with refined boundaries. Even for complicated cases, our approach still obtains accurate segmentation results.

\subsection{Ablation Studies}
In this section, we conduct our ablation studies on PASCAL VOC 2012 \emph{val} dataset, and we use the single scale results.
\begin{table}[]
\centering
\caption{Ablation study about the influence of the loss functions on PASCAL VOC 2012 \emph{val} dataset.}
\begin{tabular}{lcccc}
	\hline
	\multicolumn{2}{l}{Semantic Head} & \multicolumn{2}{l}{Feature Head} & \multirow{2}{*}{mIoU (\%)} \\ \cline{1-4}
	$\mathcal{L}_{\text{ce}}$  & $\mathcal{L}_{\text{dfr}}$  & $\mathcal{L}_{\text{fd}}$ & $\mathcal{L}_{\text{fr}}$ & \\ \hline
	\checkmark& &  &  &68.9\\
	\checkmark&\checkmark&&  &81.0\\
	\checkmark&\checkmark&\checkmark& &81.1\\
	\checkmark&\checkmark&\checkmark&\checkmark&\textbf{81.5}\\ \hline
\end{tabular}\label{tab:loss}
\end{table}

In Table~\ref{tab:loss}, we evaluate the influence of the loss functions. It can be seen that without using any loss of the feature consistency head, our proposed regularized loss brings 12.1\% mIoU increase (81.0\% \emph{v.s.} 68.9\%). Using our feature consistency head further improves the final performance, with 0.5\% mIoU growth. Besides, it can also be found that two loss functions of the feature consistency head are both useful to improve the final performance.

Table.~\ref{tab:Kernel} reports the results on applying different elements to compute the regularized kernel. By adding the deep feature, the final performance increases to 81.5\%, being 0.7\% higher than only using static information (80.8\%), which proves the effectiveness of the dynamic deep feature. It can also be found that RGB is an essential element, without it, the performance drops rapidly (from 80.8\% to 72.8\%). Nevertheless, it can also be found that adopting deep feature over spatial position can also improve the performance, with an mIoU increase of 2.1\%. Note that when deep feature is not used in $\mathcal{L}_{\text{dfr}}$ (Eq.~(\ref{eq:Kij})), we simply remove the full feature consistency head. 
It is interesting to notice that even the feature consistency head is not used, directly using deep feature in $\mathcal{L}_{\text{dfr}}$ can improve the performance (81.0\% in Table.~\ref{tab:loss} \emph{v.s.} 80.8\% in Table.~\ref{tab:Kernel}), which also proves the positive influence of the introduced deep feature. 

\begin{table}[]
	\centering
	\caption{Ablation study about the influence of the shallow feature and deep feature for our regularized loss (Eq.~(\ref{eq:Kij})) on PASCAL VOC 2012 \emph{val} dataset. ``XY" is the spatial position. ``RGB" is the color information. ``Feature" is the dynamic feature from the feature consistency head.}
	\begin{tabular}{lccc}
		\hline
		\multicolumn{3}{c}{Kernel} & \multirow{2}{*}{mIoU (\%)} \\ \cline{1-3}
		XY    & RGB    & Feature    & \\ \hline
		\checkmark& &   & 72.8\\
		\checkmark& \checkmark &   & 80.8\\
		\checkmark&        & \checkmark&74.9\\
		\checkmark& \checkmark& \checkmark &\textbf{81.5}\\ \hline
	\end{tabular}\label{tab:Kernel}
\end{table}

\begin{table}[]
	\centering
	\caption{Ablation study about the influence of the selected share feature for both semantic segmentation head and feature consistency head on PASCAL VOC 2012 \emph{val} dataset. ``Block" is shown in Fig.~\ref{fig:backbone}. }
	\begin{tabular}{lcccc}
		\hline
		\multicolumn{4}{c}{Feature} & \multirow{2}{*}{mIoU (\%)} \\ \cline{1-4}
		Block-1 & Block-2 & Block-3 & Block-4 &  \\ \hline
		\checkmark&  &  &  &81.1\\
		&\checkmark&  &  &81.1\\
		& &\checkmark& & 81.3\\
		& & &\checkmark& 81.3\\
		\checkmark& \checkmark& &  &81.2\\
		& &\checkmark&\checkmark& 80.9\\
		\checkmark& \checkmark&\checkmark&  &80.8\\
		& \checkmark&\checkmark&\checkmark& 80.5\\
		\checkmark& \checkmark& \checkmark&\checkmark&\textbf{81.5}\\ \hline
	\end{tabular}\label{tab:feature}
\end{table}

Table.~\ref{tab:feature} shows the influence of the selected feature, which is a shared feature for both the semantic segmentation head and the feature consistency head. It can be seen that the obtained performance using the feature from each block individually is sightly limited. Finally, using all features together generates the best performance. Considering that the feature map from lower block contains more low-level information and the feature map from the higher block contains more high-level information, using all of these features can supply more comprehensive representations to build accurate relationship for different pixels. 

\begin{table}[h]
	\centering
	\caption{The influence of the supervision for the feature consistency head on PASCAL VOC 2012 \emph{val} dataset. ``GT" means the provided scribble annotation. ``$M$" is our selected confident labels from the semantic segmentation head, defined in Eq.~(\ref{eq:M_i}).}
	\begin{tabular}{lcc}
		\hline
		\multicolumn{2}{l}{Supervision} & \multirow{2}{*}{mIoU (\%)} \\ \cline{1-2}
		GT & $M$ &  \\ \hline
		\checkmark& &80.6\\
		&\checkmark& \textbf{81.5}\\
		\checkmark & \checkmark& 81.1\\ \hline
	\end{tabular}\label{tab:GTM}
\end{table}

In Table.~\ref{tab:GTM}, we explore the influence of different supervision for the feature consistency head. It can be found that if only the ground truth scribble annotations are used as supervision, the performance is limited (only 80.6\%) since the ground truth can only provide limited annotations (about 3\% pixels are labeled), thus a local window will receive very few negative labels, which is insufficient for the feature distance loss $\mathcal{L}_{\text{fd}}$. Besides, using the confident prediction ($M$ defined in Eq.~(\ref{eq:M_i})) from the semantic segmentation head performs better than using both ground truth and $M$, with an mIoU gain of 0.4\% . This is because when the ground truth and $M$ are merged, it is unavoidable to introduce some incorrect pixel relationship. Specifically, there are some noisy labels in $M$, while the labels in ground truth are all correct, thus it will lead to incorrect negative pixel pairs as supervision, which is harmful for training.   

\section{Conclusion}
In this paper, we have proposed a dynamic feature regularized loss for weakly supervised semantic segmentation with scribble annotation. Our regularized loss makes full use of the static shallow feature and dynamic deep feature to build the regularized kernel, which is more accurate to describe relationship of different pixels. Meanwhile, in order to provide more powerful deep features, we introduce vision transformer as the backbone and design a feature consistency head to restrict the pair-wise pixel relationship under the supervision of the prediction from the semantic segmentation head. We found that both our regularized loss and the feature consistency head can benefit from each other and lead to a better performance. Extensive experiments show that our approach achieves new state-of-the-art performances with large margins. In the future, we plan to apply our approach on other weakly supervised semantic segmentation tasks.

\bibliographystyle{IEEEtran}
\bibliography{mybib}

\end{document}